\documentclass[10pt,twocolumn,letterpaper]{article}

\usepackage{cvpr}              % To produce the CAMERA-READY version
\definecolor{cvprblue}{rgb}{0.21,0.49,0.74}
\usepackage[pagebackref,breaklinks,colorlinks,allcolors=cvprblue]{hyperref}

\usepackage{xcolor}
\usepackage{nicematrix}
\usepackage{graphicx}
\usepackage{multirow}
\usepackage{tabularx}
\usepackage{amssymb}%
\usepackage{comment}
\usepackage{bbm}
\usepackage{tikz}
\usepackage{soul}
\usepackage[normalem]{ulem}
\usepackage{pifont}% http://ctan.org/pkg/pifont

\usepackage{xcolor} % in your preamble
\definecolor{mplblue}{RGB}{31,119,180}
\definecolor{mplorange}{RGB}{255, 165,0}
\definecolor{mplgreen}{HTML}{15b01a}
\definecolor{mplred}{HTML}{e50000}
\definecolor{mplpurple}{HTML}{7e1e9c}

\definecolor{mypurple}{HTML}{9900ff}

\definecolor{radargreen}{HTML}{d42cea}
\definecolor{radarpurple}{HTML}{1aaf6c}

\definecolor{mySquare}{rgb}{0.749,0.749,1}

\newcommand{\transsquare}{%
  \begingroup
  \setlength{\fboxsep}{0pt}%
  \colorbox{mySquare!40}{\strut\hspace{0.5em}\hspace{0.5em}}%
  \endgroup
}

\definecolor{mySquareB}{rgb}{1,0.910,0.749}
\newcommand{\transsquareB}{%
  \begingroup
  \setlength{\fboxsep}{0pt}%
  \colorbox{mySquareB!40}{\strut\hspace{0.5em}\hspace{0.5em}}%
  \endgroup
}

\def\paperID{28961} % *** Enter the Paper ID here
\def\confName{CVPR}
\def\confYear{2026}

\title{Cross-modal Counterfactual Explanations: Uncovering Decision Factors and Dataset Biases in Subjective Classification}

\author{Alina Elena Baia,
Andrea Cavallaro\\
EPFL, Lausanne, Switzerland
\\
{\tt\small \{alina.baia, andrea.cavallaro\}@epfl.ch }
}

\begin{document}

\maketitle

\begin{abstract}
Concept-driven counterfactuals explain decisions of classifiers by altering the model predictions through semantic changes. In this paper, we present a novel approach that leverages cross-modal decompositionality and image-specific concepts to create counterfactual scenarios expressed in natural language. We apply the proposed interpretability framework, termed Decompose and Explain (DeX), to the challenging domain of image privacy decisions, which are contextual and subjective. This application enables the quantification of the differential contributions of key scene elements to the model prediction. We identify relevant decision factors via a multi-criterion selection mechanism that considers both image similarity for minimal perturbations and decision confidence to prioritize impactful changes. This approach evaluates and compares diverse explanations, and assesses the interdependency and mutual influence among explanatory properties. By leveraging image-specific concepts, DeX generates image-grounded, sparse explanations, yielding significant improvements over the state of the art. Importantly, DeX operates as a training-free framework, offering high flexibility. Results show that DeX not only uncovers the principal contributing factors influencing subjective decisions, but also identifies underlying dataset biases allowing for targeted mitigation strategies to improve fairness.
\end{abstract}

\section{Introduction}
\label{sec:intro}

Language communicates more effectively than images the risks of sharing pictures containing private information~\cite{ferrarello_}.
To produce effective text-based explanations, it is necessary to establish a mapping between visual and textual modalities, enabling a model to articulate its perception of an image in an accessible, human-readable format.

Contextual text embeddings of composite concepts can be approximated as linear combinations of vector representations of their constituent factors~\cite{trager2023linear}.
Large-scale image-text multimodal embeddings exhibit linear compositionality~\cite{jia2021scaling} similar to word embeddings~\cite{fournier-etal-2020-analogies, mikolov2013efficient, logeswaran2018an}. 
This emergent arithmetic property extends across modalities in joint representation models~\cite{girdhar2023imagebind,wang2024omnibind}. Such alignment enables simple arithmetic operations on embeddings 
(e.g.~adding or subtracting text and image vectors) to perform cross-modal tasks without additional training.

In this paper, we propose DeX (Decompose and Explain), a training-free framework that utilizes cross-modal compositionality to provide text-based counterfactual explanations for image privacy classification. DeX decomposes images into key semantic elements expressed as tags, generating plausible alternative scenarios via cross-modal arithmetic in latent space\footnote{While diffusion models produce visual explanations, they necessitate guided editing and supplementary techniques (e.g.~textual inversion) which impose substantial burdens related to training, optimization, and tuning~\cite{jeanneret2024time}. DeX instead leverages the arithmetic properties of multimodal embeddings to generate explanations directly, eliminating these overheads.}. DeX evaluates the impact of these scene composition changes on privacy predictions, quantifying the relevance of image elements based on a set of desirable properties.  Our main contributions are:
\begin{itemize}
    \item Introducing a novel cross-modal (de)compositionality strategy for interpreting privacy classifier decisions. 
    \item A training-free framework that generates text-based counterfactual explanations satisfying key properties: feasibility, sparsity, and validity.
    \item A multi-criterion assessment of explanations, incorporating previously neglected aspects such as validity, confidence, concept groundedness (feasibility), and similarity.
    \item The application of DeX to analyze dataset content, revealing a significant thematic bias in a widely used 
    dataset. 
\end{itemize}

\section{Related work}

Counterfactual Explanations (CEs)~\cite{wachter2017counterfactual} define the minimal semantic changes~\cite{jacob2022steex} that flip a model's prediction~\cite{rodriguez2021beyond, dime, jeanneret2023ACE, jeanneret2024time}. 
%Since pixel-space perturbations resamble adversarial noise~\cite{Velazquez2023EvaluatingVC, GoodfellowSS14} and lack interpretability, 
%
Since pixel-based explanations (e.g.~heatmaps) lack informativeness and are difficult to interpret by non-expert users~\cite{help_me_help_the_ai, coco-cf}, recent work focuses on generating meaningful changes within the latent spaces of generative models~\cite{zemni2023octet, khorram2022cycle, jacob2022steex, stylex2021}, including VAEs~\cite{rodriguez2021beyond} and diffusion models~\cite{jeanneret2023ACE, jeanneret2024time, augustin2022diffusion}. Notably, ACE~\cite{jeanneret2023ACE} employs diffusion models to convert adversarial noise into semantic perturbations. TIME~\cite{jeanneret2024time} guides diffusion-based counterfactual generation using textual inversion to learn context and class tokens. StylEx~\cite{stylex2021} achieves instance-level explanations by training a classifier-specific StyleGAN2~\cite{stylegan2}, manipulating its StyleSpace~\cite{StyleAnalysis}. StylEx's main drawbacks are the high computational cost of per-classifier GAN training and the reliance on manual concept annotation. Other methods~\cite{ZOOM, UMO} employ pre-trained generators for model diagnosis. ZOOM~\cite{ZOOM} uses CLIP-guided text prompt differences to steer image generation, whereas UMO~\cite{UMO} finds and labels influential edit directions. A training-free alternative, DiffEx~\cite{DiffEx}, leverages VLMs to extract hierarchical domain attributes, which then guide off-the-shelf diffusion models to produce counterfactuals. However, such methods face several challenges due to the limitations of generative models, including computational cost, visual artifacts, generation failures (e.g.~missing objects), and inherited  biases~\cite{lance, coco-cf, safetypairs}. %social biases and stereotypes.
{\em DeX departs from generative approaches by operating on multimodal embeddings to apply image-representation modifications. This choice eliminates the computational overhead and biases associated with image rendering via generative models.}

Models trained with natural language latent topic descriptions~\cite{peak} or through the use of Concept Bottleneck Models (CBMs)~\cite{baia2024image} may enhance interpretability. However, they tend to underperform end-to-end counterparts~\cite{baia2024image}. 
%Additionally, the semantic grounding of the associated linear probing weights is unclear~\cite{help_me_help_the_ai}.  
{\em In contrast, DeX operates post-hoc, ensuring that the classifier's performance is unaffected. This is an advantage in high-stakes tasks such as privacy classification.}

%Originally proposed by Wachter et al.~\cite{wachter2017counterfactual} for models handling tabular data, the CE approach was subsequently adopted for vision models~\cite{rodriguez2021beyond, dime, jeanneret2023ACE, jeanneret2024time}.
%
% Because perturbations in the pixel space resemble adversarial noise~\cite{Velazquez2023EvaluatingVC, GoodfellowSS14} that offer no valuable insights, works exploit the latent space of generative network~\cite{zemni2023octet, khorram2022cycle,jacob2022steex, stylex2021}, variational autoencoder~\cite{rodriguez2021beyond}, or diffusion models~\cite{jeanneret2023ACE, jeanneret2024time, augustin2022diffusion} to produce more meaningful changes. 
% %
%

Concept Activation Vectors (CAVs) quantify the global importance of a concept by defining a corresponding direction in the model's space, derived from human-annotated positive and negative examples~\cite{kim2018interpretability}. A CAV is orthogonal to the decision boundary of a linear classifier trained on the activations of such examples. However, a major drawback is the reliance on costly human supervision and susceptibility to annotation bias. An unsupervised variant mitigates this by generating concepts via multi-resolution image segmentation and clustering~\cite{ghorbani2019towards}.
% The above methods offer global overview that explains an entire class rather than explaining individual predictions. This can be limiting in cases where a class, such as the private class in our context, contains diverse type of image content, making such explanations less informative for specific inputs. 
{\em DeX provides instead instance-specific explanations that pinpoint the concepts driving the classification decision for each individual image.}

Concept directions (CAVs) extracted from image sets~\cite{abid2022meaningfully, Akula_Wang_Zhu_2020_CoCoX} or text prompts~\cite{CounTEX} can be used as generation-free methods to produce CEs. The Conceptual Counterfactual Explanations (CCEs)~\cite{abid2022meaningfully} method learns scores to minimally adjust a sample's embedding along concept directions to alter the prediction. The minimal critical concept subset whose inclusion or removal significantly impacts the classification can also be identified~\cite{Akula_Wang_Zhu_2020_CoCoX}. CounTEX~\cite{CounTEX} derives concept directions in the CLIP space using the vector difference between the embeddings of neutral (i.e.~lacking any concepts) and concept-specific textual prompts~\cite{gal2022stylegan, kim2022diffusionclip}. As in~\cite{abid2022meaningfully}, image embeddings are counterfactually perturbed via a weighted sum of Concept Activation Vectors (CAVs) to modify the prediction. While successful on structured datasets (e.g.~AwA2~\cite{awa2_dataset}, CUB~\cite{cub_dataset}), CounTEX~\cite{CounTEX} fails to yield sparse, feasible, and diverse explanations on complex, unstructured privacy datasets~\cite{PrivacyAlert, orekondy_68_attributes}. The quality depends on the predefined concept set and optimization, leading to issues like explanation collapse and non-sparse perturbations along irrelevant concepts.
{\em DeX distinguishes itself from set-based concept methods~\cite{abid2022meaningfully, coco-cf} by operating solely on the input image. Furthermore, to perturb the embedding, DeX overcomes the limitations of CounTEX~\cite{CounTEX} by employing automatically extracted, image-specific concepts, rather than a fixed, generic set. This ensures explanations are image-grounded, highly feasible, and with improved inter-image diversity.}

%%%%%%%%%%%%%%%%%%%%%%
\section{Problem definition}
\label{sec:problem_defition}

Let $\mathcal{D} =\{(I, y)\}$
be a set of image-label pairs, where label 
$y \in \mathcal{Y}=\{pr, pu\}$ defines whether image $I$ is {\em pr}ivate or {\em pu}blic. Let $E_I$ be an image encoder and $E_T$ be a text encoder that embed their modalities into a joint multimodal space. Let $x = E_{I}(I) \in \mathbb{R}^d$ be the embedding of image $I$ and let $f: \mathbb{R}^d \rightarrow \mathbb{R}$ be a
a privacy classifier that maps the image embedding $x$ into a prediction $\tilde{y}\in \mathcal{Y}$. 

For each image $I$, $\mathcal{C}_{I} = \{c_{1}, \dots, c_{k}\}$ is a set of  
image-specific concepts expressed in natural language.  The concept types and $k$, the cardinality of $\mathcal{C}_{I}$, depend on the image content. 
The aim is to assess the impact of a missing concept, $c_{i} \in \mathcal{C}_I$, from $I$ to explain the privacy decision,  $\tilde{y} = pr$, by exploring (alternative) counterfactual scenarios. We aim to modify the embedding of image $I$ based on its concept $c_{i}$ and find a counterfactual explanation (CE) $\hat{x}$, 
such that the classifier's output $f(\hat{x})$ differs from the original prediction $\tilde{y}$: $f(\hat{x}) \neq f(x)$. If removing $c_{i}$ alters the prediction, then it constitutes a CE.

Evaluating CEs is difficult because of the scarcity of appropriate concept-based datasets and the necessity to balance multiple, often competing desiderata and associated trade-offs~\cite{wachter2017counterfactual, rodriguez2021beyond, DiCE, jeanneret2023ACE, jeanneret2024time, dime, russell2019efficient, riccardo_guidotti, verma2024counterfactual}.
In this work, we interpret feasibility as groundedness: a CE is {\em feasible} if it is grounded on the image. A CE is {\em valid} if it changes the classification outcome~\cite{wachter2017counterfactual, russell2019efficient, DiCE, jeanneret2024time, rodriguez2021beyond, CounTEX, jeanneret2023ACE}, {\em sparse} if it modifies a minimal number of attributes or features~\cite{DiCE, russell2019efficient, rodriguez2021beyond, jeanneret2024time, dime, jeanneret2023ACE, CounTEX, abid2022meaningfully}, {\em proximal} if the image (embedding) remains similar\footnote{Sparsity and proximity minimize the changes between the original embedding and its counterfactual: a sparse counterfactual (i.e. changing only a few features) will often be similar to the original instance. However, a counterfactual can be sparse yet dissimilar if changes are too large, or close in similarity but not sparse if many small modifications are made.  Hence, evaluating both offers insights into the explanation methods.} to the original~\cite{rodriguez2021beyond, CounTEX, DiCE, jeanneret2024time, dime, jeanneret2023ACE},  
{\em diverse} if explanations are different~\cite{russell2019efficient, rodriguez2021beyond, DiCE, jeanneret2023ACE, dime}.
%The CE method should generate diverse explanations to provide different viable ways of changing the model’s prediction
Additionally, we propose evaluating the counterfactual \textit{confidence}, which quantifies the classifier's support for the predicted counterfactual class.

%$\hat{x}$ be the counterfactual embedding,
% Let \(\mathcal{E}^v_I = \{\hat{x}_{1}, \hat{x}_{2}, \dots, \hat{x}_{N}\}\) be the set of valid counterfactuals, i.e.~prediction-flipping embeddings, where each explanation \(\hat{x}_{i}\) is represented by a counterfactual scenario \(c_{i}\)$, \mathcal{E}^b_I$ the set of best counterfactuals selected with a multi-objective process, $\mathcal{D}_{pr}$ the set of  correctly classified private images, and $\mathbbm{1}(\cdot)$ the indicator function.
Let \(\mathcal{E}^b_I = \{\hat{x}_{1}, \hat{x}_{2}, \dots, \hat{x}_{N}\}\) be the set of best prediction-flipping counterfactuals selected through a multi-objective process, where each explanation \(\hat{x}_{i}\) is represented by a counterfactual scenario $c_{i} \in  \mathcal{C}^b_I$, with $\mathcal{C}^b_I \subseteq \mathcal{C}_I $ (see Section~\ref{sec:counterfactual_explanations} for details).
Let $\mathcal{D}_{pr}$ be the set of  correctly classified private images, and  $\mathcal{D}_{b} = \{ I \in \mathcal{D}_{pr} \mid \mathcal{E}^b_I \neq \emptyset \}$.

We evaluate {\em feasibility}, $F$, in terms of groundedness with respect to $I$ as the average proportion of tags in CEs present in $I$:
% \begin{equation}
% F = \frac{1}{\sum_{I \in \mathcal{D}_{pr}} |\mathcal{E}^v_I|} \sum_{I \in \mathcal{D}_{pr}} \sum_{c\in \mathcal{C}^v_I} \frac{ \left|\mathcal{T}(I, l(c)) \right|}{\left| l(c) \right|},  
% \end{equation}
\begin{equation}
F = \frac{1}{\sum_{I \in \mathcal{D}_{b}} |\mathcal{E}^b_I|} 
    \sum_{I \in \mathcal{D}_{b}} 
    \sum_{c \in \mathcal{C}^b_I} 
    \frac{|\theta(I, l(c))|}{|l(c)|},
\end{equation}
where $l(\cdot)$ returns the list of $t_i$ tags composing $c$, and $\theta(\cdot, \cdot)$ is an open-set image tagger that provides the list of $t_i$ tags detected in $I$.

We quantify {\em validity}, $V$, as the ratio of  correctly classified private images with at least one prediction-flipping counterfactual:
% \begin{equation} V = \frac{1}{\left |\mathcal{D}_{pr} \right |} \sum_{I \in \mathcal{D}_{pr}}^{} \mathbbm{1} \left( \mathcal{E}^v_I \neq \emptyset \right), 
% \end{equation}
\begin{equation}
V = \frac{|\mathcal{D}_{b}|}{|\mathcal{D}_{pr}|}.
\end{equation}
Larger changes induced by counterfactuals lead to higher $V$, but this conflicts with sparsity and proximity desiderata. We will discuss the results of $F$ and $V$ in Section~\ref{sec:results} as percentages.

We measure {\em sparsity}, $S$, as the average number of tags used to generate a valid counterfactual:
%
% \begin{equation}
% S = \frac{1}{\sum_{I \in \mathcal{D}_{pr}} |\mathcal{E}^v_I|} \sum_{I \in \mathcal{D}_{pr}} \sum_{\hat{x} \in \mathcal{E}^v_I} g(\hat{x}),  
% \end{equation}
%
\begin{equation}
S = \frac{1}{\sum_{I \in \mathcal{D}_{b}} |\mathcal{E}^b_I|} 
    \sum_{I \in \mathcal{D}_{b}} \sum_{\hat{x} \in \mathcal{E}^b_I} g(\hat{x}),
\end{equation}
where $g(\hat{x})$ computes the number of concepts used to generate $\hat{x}$. Enforcing sparsity may compromise validity. 
To assess {\em proximity}, $P$, we consider the average cosine similarity, $\text{cos}(\cdot)$, between the image embeddings $x$ and that of its counterfactual $\hat{x}$ over all explanations in $\mathcal{E}^b_I$:
% \begin{equation}
% P = \frac{1}{\sum_{I \in \mathcal{D}_{pr}} |\mathcal{E}^v_I|} \sum_{I \in \mathcal{D}_{pr}} \sum_{\hat{x} \in \mathcal{E}^v_I} \text{cos}(x, \hat{x}).  
% \end{equation}
\begin{equation}
P = \frac{1}{\sum_{I \in \mathcal{D}_{b}} |\mathcal{E}^b_I|} 
    \sum_{I \in \mathcal{D}_{b}} \sum_{\hat{x} \in \mathcal{E}^b_I} 
    \text{cos}(E_{I}(I), \hat{x}).
\end{equation}

The average {\em confidence}, $C$, of the new predictions in the counterfactual class is: 
% \begin{equation}
% C = \frac{1}{\sum_{I \in \mathcal{D}_{pr}} |\mathcal{E}^v_I|} \sum_{I \in \mathcal{D}_{pr}} \sum_{\hat{x} \in \mathcal{E}^v_I} p(\hat{x}),
% \end{equation}
%
\begin{equation}
C = \frac{1}{\sum_{I \in \mathcal{D}_{b}} |\mathcal{E}^b_I|} 
    \sum_{I \in \mathcal{D}_{b}} \sum_{\hat{x} \in \mathcal{E}^b_I} p(\hat{x}),
\end{equation}
where $p(\hat{x})$ is the confidence of  prediction $f(\hat{x})$. 

We measure \textit{diversity}, $D$, by computing the average across the dataset of the pairwise cosine similarity between explanations of the same $I$:
%
% \begin{equation}
% D = \frac{2}{N (N - 1)|\mathcal{D}_{v}|}  \sum_{I \in \mathcal{D}_{v}} 
% \sum_{i=1}^{N} \sum_{j=i+1}^{N} 
% \text{cos}\big(E_{T}(c_{i}), E_{T}(c_{j})\big),
% \end{equation}
%
\begin{equation}
D = \frac{1}{N (N - 1) |\mathcal{D}_{b}|}  
\sum_{I \in \mathcal{D}_{b}} 
\sum_{\substack{c_i, c_j \in \mathcal{C}^b_I \\ i < j}}
\big( 1 - \cos( E_{T}(c_i), E_{T}(c_j) ) \big),
\end{equation}
where $N  = |\mathcal{C}^{b}_I|$. Higher values of $D$ indicate higher diversity among explanations. Additionally, to ensure the avoidance of repetitive explanations across different images, we detect \textit{explanation collapse}.
For all images, we compute the centroid of their explanations:
\begin{equation}
\bar{c}_I = \frac{1}{N} \sum_{c_i \in \mathcal{C}^{b}_I} E_{T}(c_{i}),
\end{equation}
and then we compute their average pairwise similarity:
%
% \begin{equation}
% R = \frac{2}{|\mathcal{D}_{v}| (|\mathcal{D}_{v}| - 1)}
% \sum_{I \in \mathcal{D}_{v}} 
% \sum_{\substack{J \neq I \\ J \in \mathcal{D}_{v}}} \text{cos}(\bar{c}_I, \bar{c}_J).
% \end{equation}
\begin{equation}
R = \frac{1}{|\mathcal{D}_{b}| (|\mathcal{D}_{b}| - 1)}
\sum_{I \in \mathcal{D}_{b}} 
\sum_{\substack{J \neq I \\ J \in \mathcal{D}_{b}}} 
1 - \cos(\bar{c}_I, \bar{c}_J).
\end{equation}
%
%
% where $\mathcal{D}_{v} = \{ I \in \mathcal{D}_{pr} \mid \mathcal{E}^v_I \neq \emptyset \}$.
%
% Specifically, we consider each explanation to be a topic, and we compute the average pairwise cosine similarity between explanations. Given the set $\mathcal{}\mathcal{E}$ of all $N$ valid explanations across the dataset, $\mathcal{E} = \bigcup_{I \in \mathcal{D}_{pr}} \mathcal{E}^v_I= \{\hat{x}_1, \hat{x}_2, \dots, \hat{x}_N\}$, where each $\hat{x}_i$ is represented by a counterfactual scenario $c_i$, we compute the inter-topic similarity ($R$) as:
% \begin{equation}
% ITS  = \frac{2}{N(N-1)} \sum_{i=1}^{N} \sum_{j=i+1}^{N}\text{cos}(E_{txt}(c_i), E_{txt}(c_j)).   
% \end{equation}
%
Higher values of $R$ indicate greater diversity of explanations across images, corresponding to more image-specific and contextually relevant explanations.
%

% 

% \noindent\textbf{NOTE TO self: not really part of initial list of quality criteria. Metrics wise, looking at the results that I had in the table, does not give me improvements, except for the ITS metrics, so we might consider to remove this to give more visibility to the other results.}

%%%%%%%%%%%%%%%%%%%%%%%%%%%%%%%%%%%%%%%
\section{Decompose and Explain (DeX)}

%First, we check this approach (Sec 4.1), then we detail how it is implemented (Sec 4.2).  

\subsection{Cross-modal arithmetic}
\label{sec:cross_modal_section}
% NOTE: ADD FIGURE WITH VECTORS ILLUSTRATING THE KEY IDEA.
\begin{figure*}[t]
    \centering
    \includegraphics[width=\textwidth]{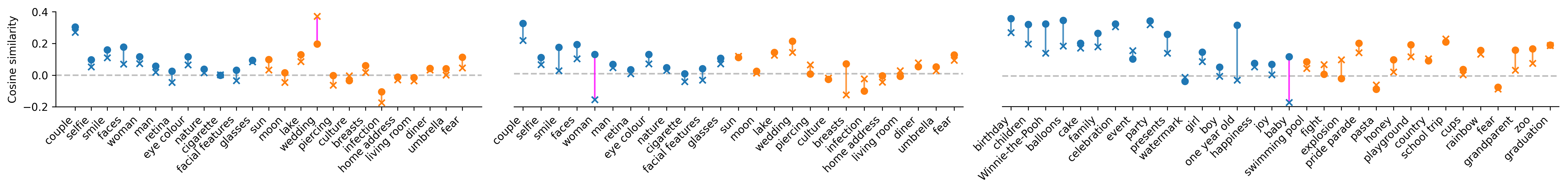}

    \caption{Examples of cross-modal arithmetic. Concept addition (left) selectively increases target similarity while preserving overall similarity to others. Concept removal (middle, right) results in a localized reduction in similarity for the target and semantically related concepts. Key- {\color{black}\large$\bullet$}/{\color{black}\large{$\times$}}: cosine similarity values before/after the arithmetic;{\color{mplblue}\rule[0.5ex]{1em}{0.8pt}}/{\color{mplorange}\rule[0.5ex]{1em}{0.8pt}}: related/unrelated concepts; {\color{magenta}\rule[0.5ex]{1em}{0.8pt}}: edited concept. } 
    \label{fig:cross_modal_arith}
\end{figure*}

We investigate the extent to which the embedding space supports semantic composition by observing how the embeddings transform when concepts are arithmetically added.
Similar to~\cite{splice}, we performed a linearity test to check that concatenating two inputs results in an embedding similar to the sum of the individual embeddings. 
We considered the text modality: given concepts $c_a$ and $c_b$, we append them to form $c_{ab} = "c_a, c_b"$, embed each of $c_a$, $c_b$, and $c_{ab}$  using OmniBind~\cite{wang2024omnibind} ($E$), and compute the cosine similarity ($CS$) between $E(c_{ab})$ and $E(c_a)+E(c_b)$.  Across 1000 randomly selected $c_a$, $c_b$ pairs and 3 runs, the mean (std) $CS$ is 0.61 (0.13). For reference, the mean pairwise $CS$ between synonyms of “car” taken from an online \href{https://www.powerthesaurus.org/}{thesaurus} is 0.35.
We extended the analysis to triplets of concepts, computing the $CS$ of the concatenated triplet embedding and the sum of individual embeddings.
Results show a mean (std) $CS$ of  0.53 (0.11). These results suggest that OmniBind embeddings exhibit linear compositionality.

%

%

% Moreover, we use an image retrieval-based approach to visually examine how the  embeddings change when concepts are added or removed via arithmetic. The goal is to understand whether the embedding space supports semantic composition (e.g., removing "car" from should reduce the retrieval of car-related images). 
% We manually analyze a few cases. Starting from captions created for images, we modify the caption embeddings by subtracting or adding the embedding of a concept. We then retrieve images from an image database using both the original and modified caption embeddings, allowing us to observe the effect of the embedding manipulation.
% %
% Examples are shown in Figure~\ref{fig:emb_arith_txt_txt}. Note that due to the limited size of the image dataset used for retrieval (VISPR dataset, 10k samples),  we expect cases where retrieved images may not perfectly align with the caption. 
% %
% The results 
% of image retrieval show 
% that the modified embeddings retrieves images that align with the absence of removed concepts, or the combined meaning of added concepts. For example, removing \textit{a woman }from \textit{a woman in a green field} retrieves an image of an empty field, while removing \textit{car} from \textit{a red car} retrieves a red object (rose) without a car. Removing \textit{performing surgery} from \textit{doctors performing surgery } retrieves people sitting casually, reflecting the removal of the surgical activity. 
% This indicates that the embedding space supports semantic composition between concepts.

Next, we generate modified caption embeddings by arithmetically adjusting them with a concept embedding, and compare the subsequent image retrieval results to the original ones to confirm the intended semantic alteration.  For instance, removing \textit{a woman} from \textit{a woman in a green field} retrieves images of empty fields; removing \textit{car} from \textit{a red car} retrieves a red object without a car; subtracting \textit{performing surgery} from \textit{doctors performing surgery} returns casual scenes without the surgical context. This confirms that the embedding space supports the semantic composition of concepts.

We also investigate cross-modal composition via embedding arithmetic by examining how image–concept similarities change when new concepts are added or existing concepts are removed from an image (see Figure~\ref{fig:cross_modal_arith}). 
%Adding the text embedding of a new concept to an image embedding should enrich the image representation to reflect the new concept, while retaining the similarity with concepts originally present in the image.  
%
%Subtracting the embedding of an existing concept from an image should reduce similarity to that concept while maintaining similarity with other existing concepts.
%
We randomly sampled images and defined for each image a set of relevant and irrelevant concepts. We first establish a baseline by computing the cosine similarity between the original image and both concept sets. We hypothesize high similarity for relevant concepts and low similarity for irrelevant ones. We then recompute concept similarities: adding an irrelevant concept should increase similarity with the added concept while preserving similarity with the original relevant ones; removing a relevant concept should decrease similarity with that concept while maintaining similarity with the others.
%
% We add or subtract the embeddings of concepts from the original image and recompute the similarity between the modified image embeddings and the concepts. We expect increased similarity with the added concept while preserving similar similarity with the original relevant concepts. For the removal we expect decreased similarity with the removed concepts while  maintaining similarity with the other relevant concepts of the original image. 
%
When adding text embeddings of new concepts to image embeddings, the cosine similarity with the added concept increased significantly, while maintaining similarity with pre-existing, related concepts. For example, adding \textit{wedding} to an initially unrelated image resulted in a higher similarity to \textit{wedding}, with minimal variations to other original concepts. %Similarity with certain related concepts also increased (e.g., festival ),  potentially due to contextual associations (e.g., festivals involve people). 
Overall, adding a new concept increases similarity with that concept while generally preserving similarity with existing related ones.
Furthermore, when a concept is removed (e.g., \textit{baby}), the similarity between the modified image representation with that concept significantly decreases, as well as the similarity with related concepts (e.g., \textit{one year old}). Similarities with unrelated concepts remain generally stable, indicating that the manipulation affects the removed concept and its semantically related concepts without impacting other concepts.

% When removing a concept (i.e. bud from a picture of an orange), the cosine similarity of the modified image with the removed concept significantly decreases, as well as the similarity with elements that are related to the removed concept (i.e. flower stalk, stem, fruit).

\subsection{Counterfactual explanations}
\label{sec:counterfactual_explanations}

\begin{figure}[t!]
    \centering
   \includegraphics[width=1\columnwidth]{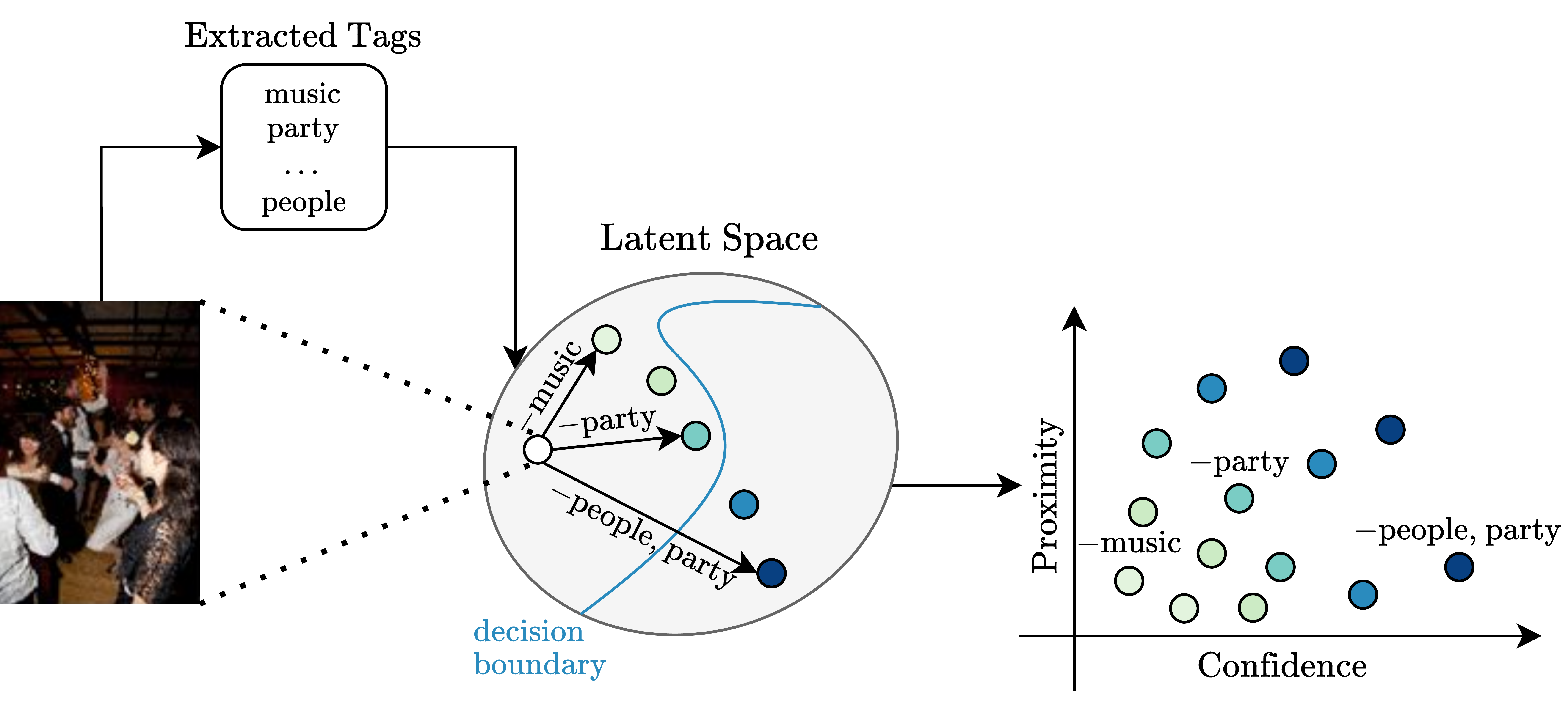}

    \caption{DeX generates concept-based explanations for a given image $I$ via a 3-step process: concept extraction and counterfactual scenario creation, cross-modal decomposition for image representation manipulation in the latent space, and  multi-criterion selection to identify privacy-relevant scenarios (i.e.~explanations).}
    \label{fig:xxx}
\end{figure}

\begin{figure}[t!]
\centering
\begin{tabular}{cc}
    % --- Top row ---
    \hspace{-0.4cm}
    \includegraphics[width=0.5\columnwidth, trim=8.2cm 0cm 0cm 0cm, clip]{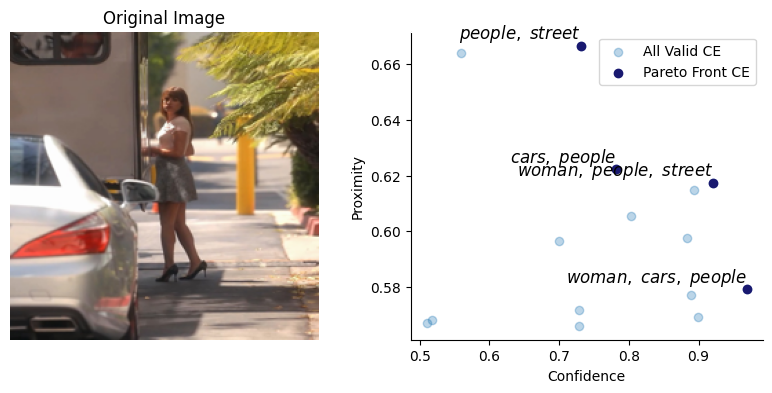}&
   \hspace{-0.4cm}
    \includegraphics[width=0.5\columnwidth, trim=8.2cm 0cm 0cm 0cm, clip]{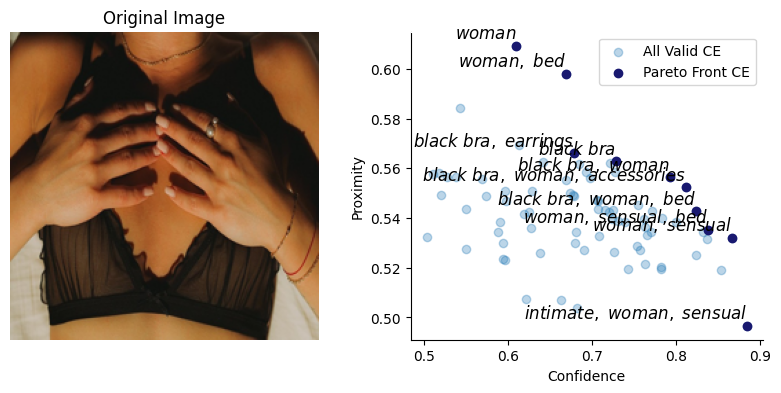}  \\

    % --- Bottom row with added horizontal spacing ---
    \hspace{0.4cm}%
      \includegraphics[width=0.4\columnwidth]{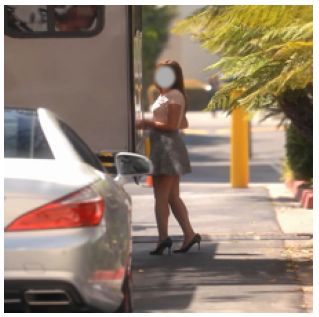} &
      \hspace{0.4cm}%
    \includegraphics[width=0.4\columnwidth]{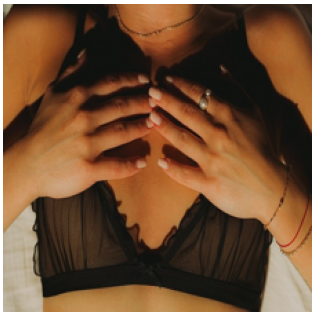} \\

\end{tabular}

\caption{Confidence-proximity Pareto trade-off of counterfactual explanations (top) and the original images (bottom): it shows the interaction between competing criteria and how different concepts influence the model's decision-making.  }
\label{fig:paretos}
\end{figure}

\begin{figure*}[t]
    \centering
       \includegraphics[width=1\textwidth]{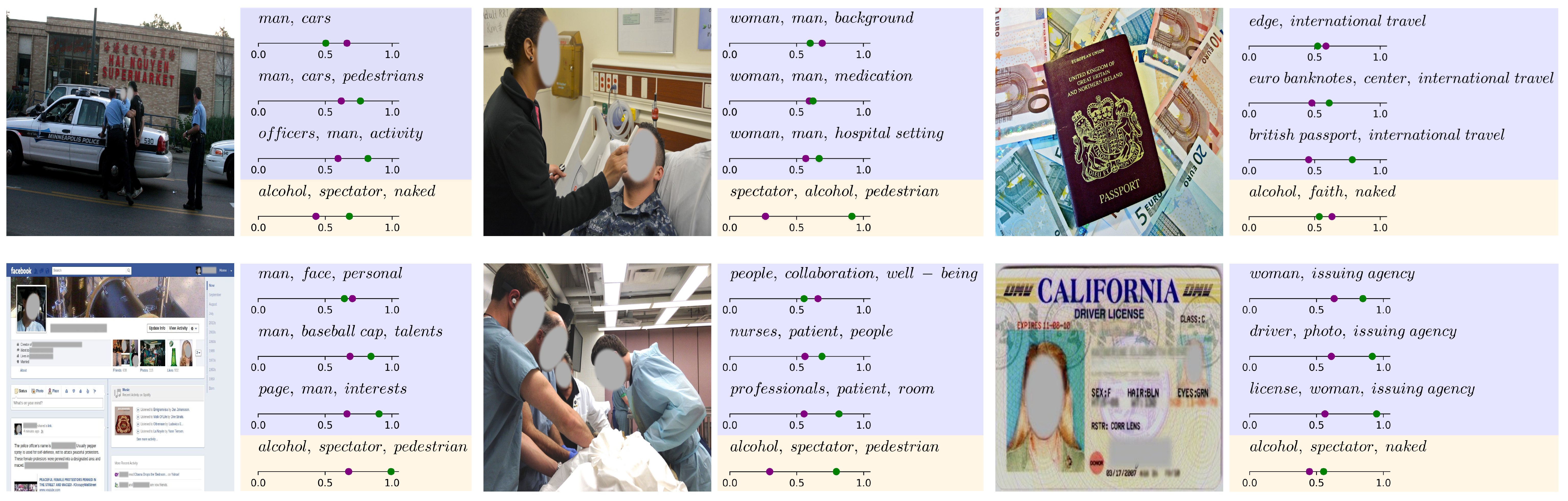}

    \caption{ Sample explanations by DeX  (\transsquare) and CounTEX (\transsquareB). Criteria: confidence ($C$,  {\color{mplgreen}\large$\bullet$}) and proximity ($P$, {\color{mplpurple}\large$\bullet$}) via cosine similarity. Note that the explanations by CounTEX~\cite{CounTEX} (\transsquareB) are repetitive and not grounded in the image.}
    \label{fig:examples_of_ce}
\end{figure*}

Based on the problem definition in Section~\ref{sec:problem_defition} and the validation of the properties in Section~\ref{sec:cross_modal_section}, we propose DeX, a training-free framework based on cross-modal compositionality to provide text-based counterfactual explanations. DeX first generates a set of counterfactual scenarios with grounded concepts and then probes each scenario using embedding arithmetic, checking the model's prediction against the resulting manipulated embedding representation (see Figure~\ref{fig:xxx}). 

To obtain the set of candidate scenarios $\mathcal{C}_I$, DeX generates an image description $d$ for image $I$ through an instruction $p$ (e.g.~\textit{Describe this image as detailed as possible}) via 
an instruction following 
Large Vision-Language Model (LVLM) designed to generate text from an input image (InstructBLIP~\cite{instructblip}).
The description captures objects, their attributes, and image context (e.g.~\textit{the image depicts a romantic moment between a man and a woman}). DeX then generates the tag-based summarization of $d$, $\mathcal{T}_I = \{t_1, \cdots, t_n \}$ (e.g.~\textit{man, woman, romantic moment}), through an LLM (Vicuna~\cite{vicuna2023}).

We derive $\mathcal{C}_I$ from $\mathcal{T}_I$ by generating all unique combinations of elements in $\mathcal{T}_I$ up to a specified length $s$ as:
\begin{equation}
\mathcal{C}_I = \bigcup_{i=1}^{s} \binom{\mathcal{T}_I}{i} = \bigcup_{i=1}^{s} \{ S \subseteq \mathcal{T}_I, \text{ s.t. } |S| = i\},
\end{equation}
where $i\in \mathbb{N}$ (e.g.~a scenario of length 2 is \textit{woman, romantic moment}). To enforce sparsity, for DeX we set $s = 3$, as previous research~\cite{MILLER20191} showed that people prefer concise explanations.

Given a correctly-classified pair $(I,pr)$, we compute a candidate CE $\hat{x}_j$ for $I$  by removing $c_j\in \mathcal{C}_I$ from its representation. Specifically, following~\cite{CounTEX}, for each concept $c_j$, we define its direction $e_{c_j}$ in the latent space as the difference between a concept-specific text prompt ($t_{\text{trg}}$)  and a concept-neutral anchor prompt ($t_{\text{src}}$, e.g.~\textit{a photo of object}): $e_{c_j} = E_T(t_{\text{trg}}) - E_T(t_{\text{src}})$. The resulting vector is then normalized to unit length.
The candidate CE $\hat{x}_j$ is obtained as $\hat{x}_j =  x -  e_{c_j}$ and 
the set of all candidate CEs for $I$ is $\mathcal{E}_{I} = \{ \hat{x}_j| j = 1, \dots, k\}$.

Next,  we evaluate the classifier predictions for each $\hat{x}_j$. We define the set of all  valid image-based CEs for $I$ as 
$\mathcal{E}^{v}_{I} \subseteq \mathcal{E}_{I}$ such that $\hat{x}_j \in \mathcal{E}^v_I \iff f(\hat{x}_j)  \neq f(x)$,
and their corresponding set of $c_j$ concepts as  $\mathcal{C}^v_I$.
As $c_j$ is used to generate $\hat{x}_j$, we refer to $\hat{x}_j$ (image-based) and $c_j$ (textual) as CEs interchangeably. Figure~\ref{fig:paretos} shows the confidence-proximity tradeoff for valid counterfactuals with Pareto-optimal explanations highlighted.

DeX generates a set of image counterfactuals $\mathcal{E}^v_I$. 
Selecting the optimal set of CEs involves accounting for multiple conflicting criteria such as prediction confidence and proximity. We adopt a multi-objective optimization approach that identifies the Pareto front, a set of solutions that are non-dominated with respect to all criteria, i.e.~no other solution performs better in terms of all criteria simultaneously.

Let each $c \in \mathcal{C}^v_I$ be associated with a vector-valued objective function  $\mathbf{o}(c) = \big(o_1(c), o_2(c), ..., o_m(c) \big)$, where $o_i$ 
represents an objective function (i.e.~prediction confidence, proximity), $i = 1, \dots m$, and $m \geq 2$. We define the dominance relationships from a maximization perspective: a counterfactual solution $c_j$ dominates a solution $c_z$ (denoted as $c_j \prec c_z$) if $o_i(c_j) \geq o_i(c_z), \forall i = 1, \dots, m$ and $ \exists i = 1, \dots, m$ such that $o_i(c_j) > o_i(c_z)$. 
We then obtain the Pareto front:
\begin{equation}
    \mathcal{P} = \{ c \in \mathcal{C}^v_I| \nexists c'  \in \mathcal{C}^v_I  \text{ s.t. } c' \prec c \},
\end{equation} 
which may contain many optimal solutions. We then select the subset $\mathcal{C}^b_I $  of $q$ solutions from $\mathcal{P}$ 
that minimizes inter-explanation similarity:
\begin{equation}
\mathcal{C}^b_I = \underset{|S|=q, S \subseteq \mathcal{P}}{\arg\min} \sum_{c_i, c_j \in S, i \neq j} \cos(E_T{(c_i)}, E_T{(c_j)}), 
\end{equation}
This selection process ensures that $\mathcal{C}^b_I $ not only satisfies all non-dominated criteria but also maximizes the diversity among the selected counterfactual explanations. $\mathcal{E}^b_I $ is the image-based set of CEs  corresponding  to $\mathcal{C}^b_I $. We set $q =3$. 

By adopting this multi-objective optimization strategy, DeX generates high-quality CEs with various trade-offs between the desiderata (see Figure~\ref{fig:examples_of_ce}). 
For example, the privacy decision of the image depicting a driver's license can be explained with  \textit{license, woman, issuing agency} or \textit{driver, photo, issuing agency}. The first CE can be mapped to the VISPR~\cite{orekondy_68_attributes} ground-truth \textit{drivers\_license}, \textit{gender}, and \textit{address\_home\_partial}. The second CE can be linked to the ground-truth \textit{face\_complete} since it is about the photo of the driver. The decision of an image showcasing a group of medical professionals and a patient in a hospital room can be explained with \textit{people, collaboration, well-being}, with \textit{professionals, patient, room}, or with \textit{nurses, patient, people}
which can be mapped to the ground truth \textit{relantionship\_professional} and \textit{occupation}. 

Note that {feasibility} is enhanced by design, as explanations are built upon a set of image-derived concepts and hence the explanations are grounded in the image content.

%%%%%%%%%%%%%%%%%%%%%%
\section{Results}
\label{sec:results}

\noindent\textbf{Experimental setup}.  We compare  DeX with CounTEX~\cite{CounTEX}, which generates textual CEs by manipulating image embeddings via text-driven concepts, making it a direct reference to our work. Additionally, we analyze explanations through topic modeling and discuss the main factors driving privacy decisions and potential biases. 
We use a linear classifier trained on OmniBind embeddings~\cite{wang2024omnibind}, OB+linear, for both DeX and CounTEX~\cite{CounTEX} for a fair comparison.
Table~\ref{tab:privacy_classification_results} shows
the accuracy and F1-macro scores for the classification on two privacy datasets. %PrivacyAlert and VISPR datasets.
\begin{table}[t!]
\setlength\tabcolsep{5pt}
\centering
\caption{Classification performance with a linear classifier trained with OmniBind embeddings~\cite{wang2024omnibind}. Results are reported on the PrivacyAlert~\cite{PrivacyAlert} and VISPR~\cite{orekondy_68_attributes} test sets.  Key: $ACC$ (\%): accuracy, $F1$‑m (\%): F1‑macro.}
\label{tab:privacy_classification_results}
\begin{tabular}{l  c c}
\toprule
\textbf{Dataset} &   \multicolumn{1}{c}{$ACC$} & \multicolumn{1}{c}{$F1$‑m} \\
\midrule
\multirow{1}{*}{PrivacyAlert}

     & 87.17  & 83.34  \\

% \midrule
\multirow{1}{*}{VISPR}

    & 91.19 & 90.92 \\

\bottomrule
\end{tabular}
\end{table}
We trained the OB+linear classifier for 100 epochs, using the Adam optimizer (learning rate of $10^{-3}$, batch size 64) with standard cross-entropy loss. 
As CounTEX requires a predefined concept list, we adopt the privacy taxonomy from~\cite{PrivacyAlert} and we manually augment it to expand the public concepts by adding concepts describing clusters of public images~\cite{baia2024image}.
Using this predefined concept library with $L$ concept keywords, $\mathcal{C} = \{c_1, \dots, c_L \}$, CounTEX assigns a weight
$ w_c(\hat{x})$ to each concept $c$ when generating the counterfactual $\hat{x}$.  
We take the top-3 concepts with the highest negative $w_c$ (those that explain the private class) to showcase examples  of explanations, and to assess the feasibility and explanation collapse since the sparsity of this method is usually high (i.e., many concepts are used to modify the image). For computing the sparsity, we consider the concepts with weights larger than 0.1, that is $g(\hat{x}) = \left | \{c \in \mathcal{C} | w_c(\hat{x}) >0.1 \} \right | $.

As the code is not publicly available, we implement the method following the paper's instructions and setup. We optimize the concept scores $w$ using the SGD optimizer (learning rate $10^{-2}$, maximum 100 iterations) with early stopping triggered by prediction change. We initialize $w$ using Xavier initialization and use cross-entropy loss, identify loss with 0.1 regularization parameter together with $L_1$ and $L_2$ with a regularization parameter of 0.1 to optimize $w$.

To assess the explanation collapse we use as image tagger the Recognize Anything Model (RAM)~\cite{huang2023open}, a model capable of assigning multiple semantic labels to an image and generalizing to unseen categories. To obtain the Pareto front, we use the implementation available in the pymoo~\cite{pymoo} Python library. We use SBERT~\cite{sentenceTransformer} to encode the explanations and compute the similarities.

We use two publicly available datasets: PrivacyAlert~\cite{PrivacyAlert} and VISPR~\cite{orekondy_68_attributes}. 
PrivacyAlert contains more explicit content, while VISPR includes content like documents (passports, IDs, emails). The two datasets are complementary and have differing annotations for similar content (e.g., images with cars or children).

\begin{figure}[t!]
    \centering
    \includegraphics[width=0.49\columnwidth]{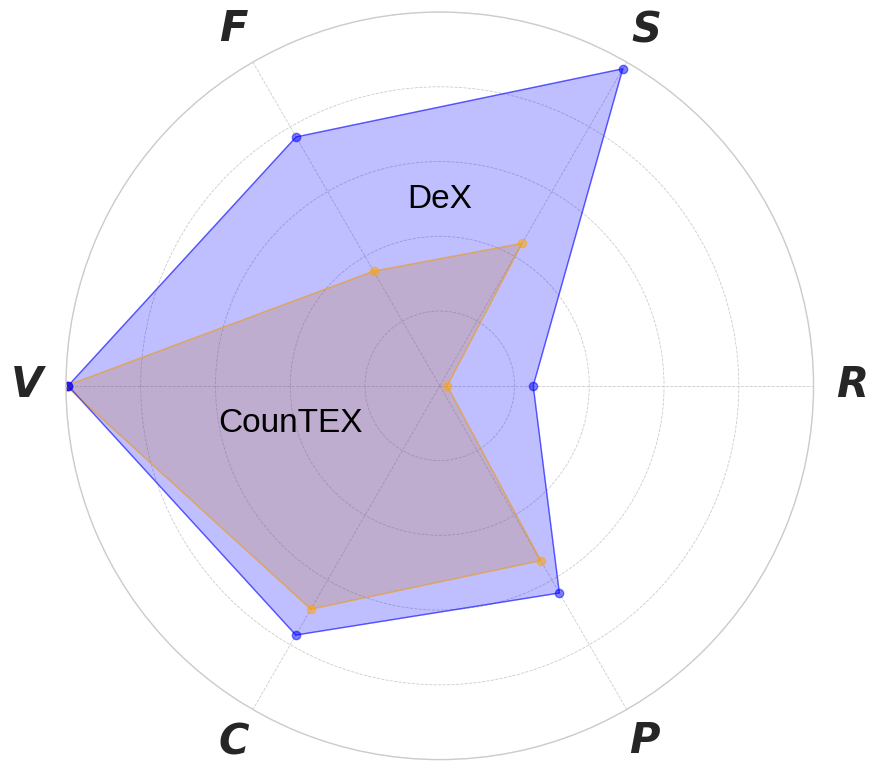}
    \includegraphics[width=0.49\columnwidth]{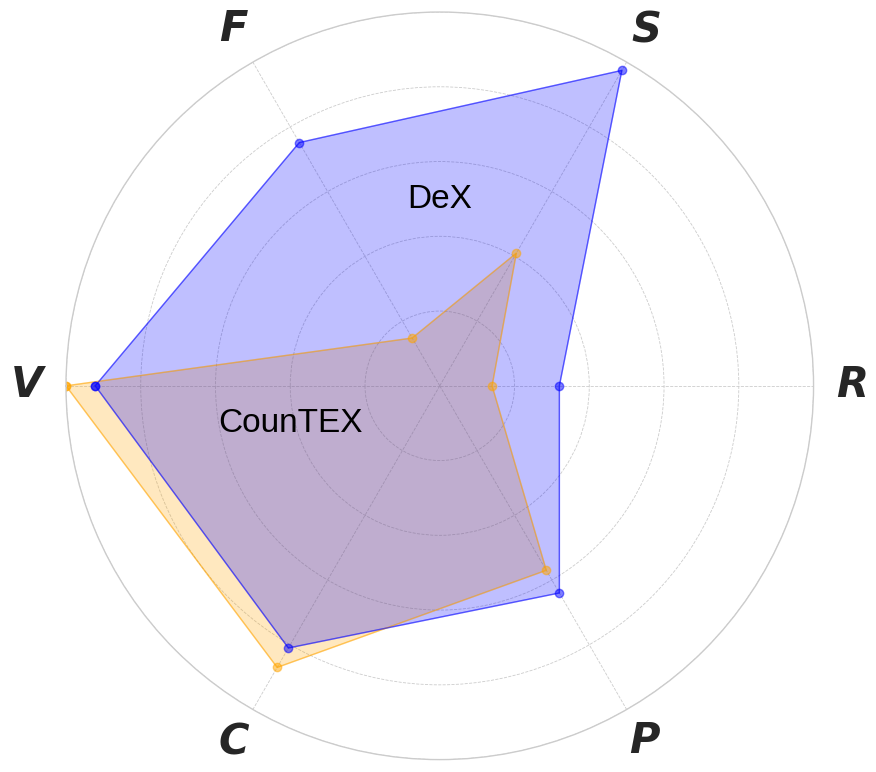}
    
    \caption{Comparison of DeX and CounTEX across validity, $V$, feasibility, $F$, sparsity, $S$, explanation collapse, $R$,  proximity via cosine similarity between the original image embeddings and its counterfactual, $P$,  and confidence, $C$,  on PrivacyAlert~\cite{PrivacyAlert} (left) and VISPR~\cite{orekondy_68_attributes} (right). 
    For visualization, $S$ is scaled to [0,1] and inverted (i.e~the higher, the better). For scaling, the maximum value of $S$ is set to 100. For CounTEX, $F$ and $R$ are reported with respect to the top-3 concepts. }
    \label{fig:radar_plots}
\end{figure}

\begin{figure}[t!]
    \centering
    \includegraphics[width=0.47\columnwidth]
        {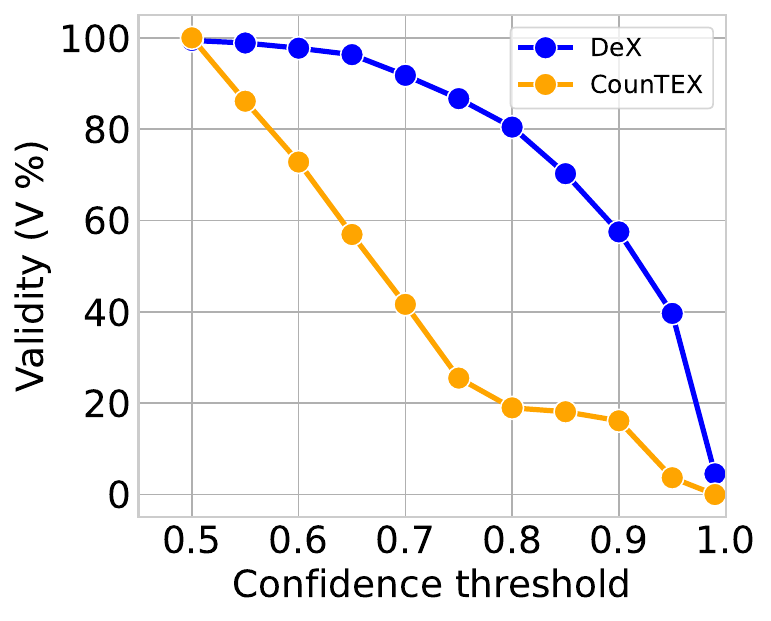}
    \includegraphics[width=0.47\columnwidth]
        {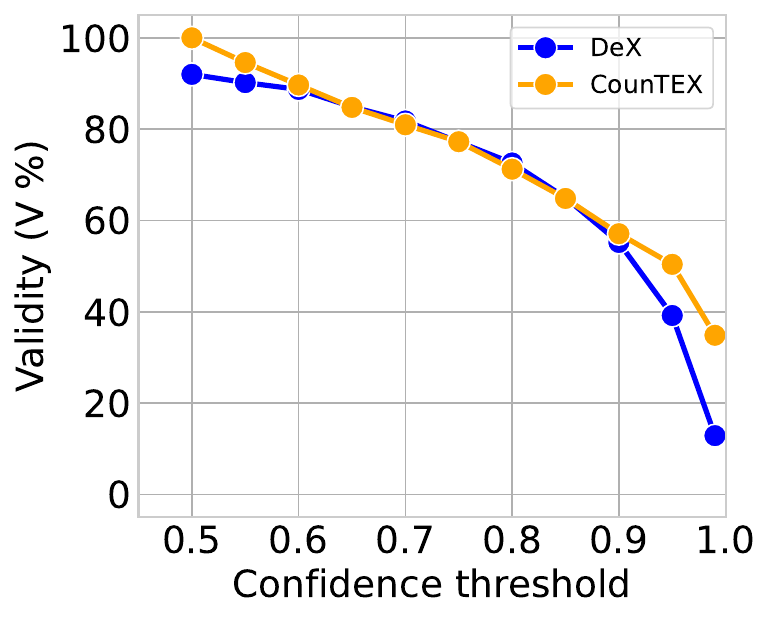}
        
    \caption{Validity at different confidence thresholds, PrivacyAlert~\cite{PrivacyAlert} (left) and VISPR~\cite{orekondy_68_attributes} (right). DeX generates high-confidence explanations with high sparsity. 
    CounTEX~\cite{CounTEX} produces comparable high-confidence explanations but with lower sparsity for VISPR, whereas it fails to produce high-confidence explanations for PrivacyAlert.   }
    \label{fig:validity_vs_confidence}
\vspace{-0.5cm}
\end{figure}

\begin{figure}[t!]
    \centering
    \includegraphics[width=0.47\columnwidth]
        {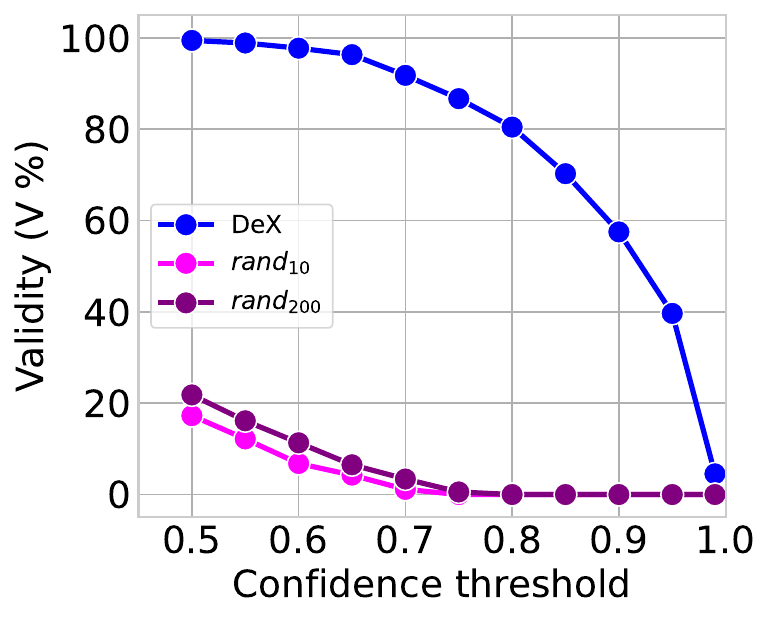}
    \includegraphics[width=0.47\columnwidth]
        {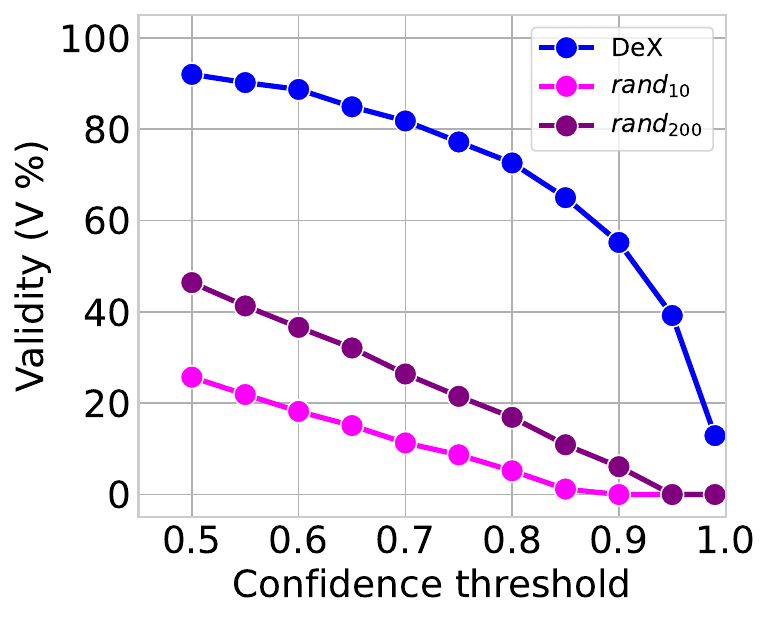}

    \caption{Robustness to perturbations,   PrivacyAlert~\cite{PrivacyAlert} (left) and VISPR~\cite{orekondy_68_attributes} (right). During the best explanation selection, DeX filters out the prediction flips caused by perturbations. }
    \label{fig:random_perturbations}
    \vspace{-0.5cm}
\end{figure}

\noindent\textbf{Discussion}.
Figure~\ref{fig:radar_plots} shows the quantitative results. For our method we report results with respect to the set  $\mathcal{E}^b_I$ of best counterfactuals selected with two criteria: prediction confidence ($C$) and proximity ($P$) via cosine similarity.
Moreover, we also report results using random perturbations to alter the image, denoted as $rand$ (Figure~\ref{fig:random_perturbations}).
Our method achieves a validity ($V$) of 99.43\% on the PrivacyAlert dataset and 92\% on the VISPR dataset while also having high sparsity. In contrast, CounTEX can reach 100\% $V$ on both datasets, but at the cost of low sparsity, resulting in image manipulations that affect many concepts, often above 50. Moreover, when high-confidence prediction is enforced CounTEX's validity drops significantly on the PrivacyAlert dataset (Figure~\ref{fig:validity_vs_confidence}), despite the high number of concepts used to manipulate the image. DeX achieves high-confidence prediction changes with high sparsity. 
In terms of confidence ($C$), DeX outperforms CounTEX on PrivacyAlert and achieves comparable results on VISPR.
The more reliable public predictions after the image perturbation indicate a better ability to remove privacy-relevant concepts.
DeX has a higher proximity ($P$) of about 0.64, compared to CounTEX’s scores, which are in the range of 0.54–0.57.
A higher $P$ score can be attributed to the fundamental differences in the concept editing process.
Our method enforces sparsity by design and limits the number of concepts used to modify the image. This differs from CounTEX, which manipulates the image with respect to many concepts which leads to reduced proximity.  

Since CounTEX generates only a single explanation per image, we compare the methods in terms of variations across the dataset ($R$) rather than focusing on intra-image diversity.
The results show that DeX achieves greater diversity in its explanations across different samples, while CounTEX suffers from explanation collapse, generating identical or highly similar explanations for a large portion of the images. This could be due to its reliance on the fixed, predefined set of concepts, which may be generic or and not well-aligned with the dataset's content. Furthermore, DeX achieves diversity ($D$) scores of 0.29 and 0.22 on the PrivacyAlert and VISPR datasets, respectively. These values suggest diversity among intra-image explanations, showing  DeX's ability to capture different relevant concepts.

DeX also achieves high $F$ scores with a consistent range of 75-77\% across both datasets, showing that the explanations are well-grounded and relevant to the input images. In contrast, CounTEX has a lower range of $F$ scores, from 12\% to 35\% across the various configurations of the number of concepts. In particular, when using only the top-3 concepts leads to a higher $F$ compared using top-10 concepts. 
For PrivacyAlert, the top-3 setting achieves an  
$F$ of 35.41\%
compared to 18.16\% 
when using the top-10 concepts. A similar trend is observed for VISPR, where the top-3 concepts reach an  
$F$ of 14.76\%, 
outperforming the top-10 setting, which attains an $F$ of 12.06\%. 
This indicates that only a small subset of concepts are grounded, leading to CounTEX's explanations that lack contextual relevance.

\begin{figure}[b]
\vspace{-0.4cm}
\centering
\begin{tabular}{@{} m{0.3\columnwidth} m{0.6\columnwidth} @{}}
  \includegraphics[width=0.32\columnwidth]{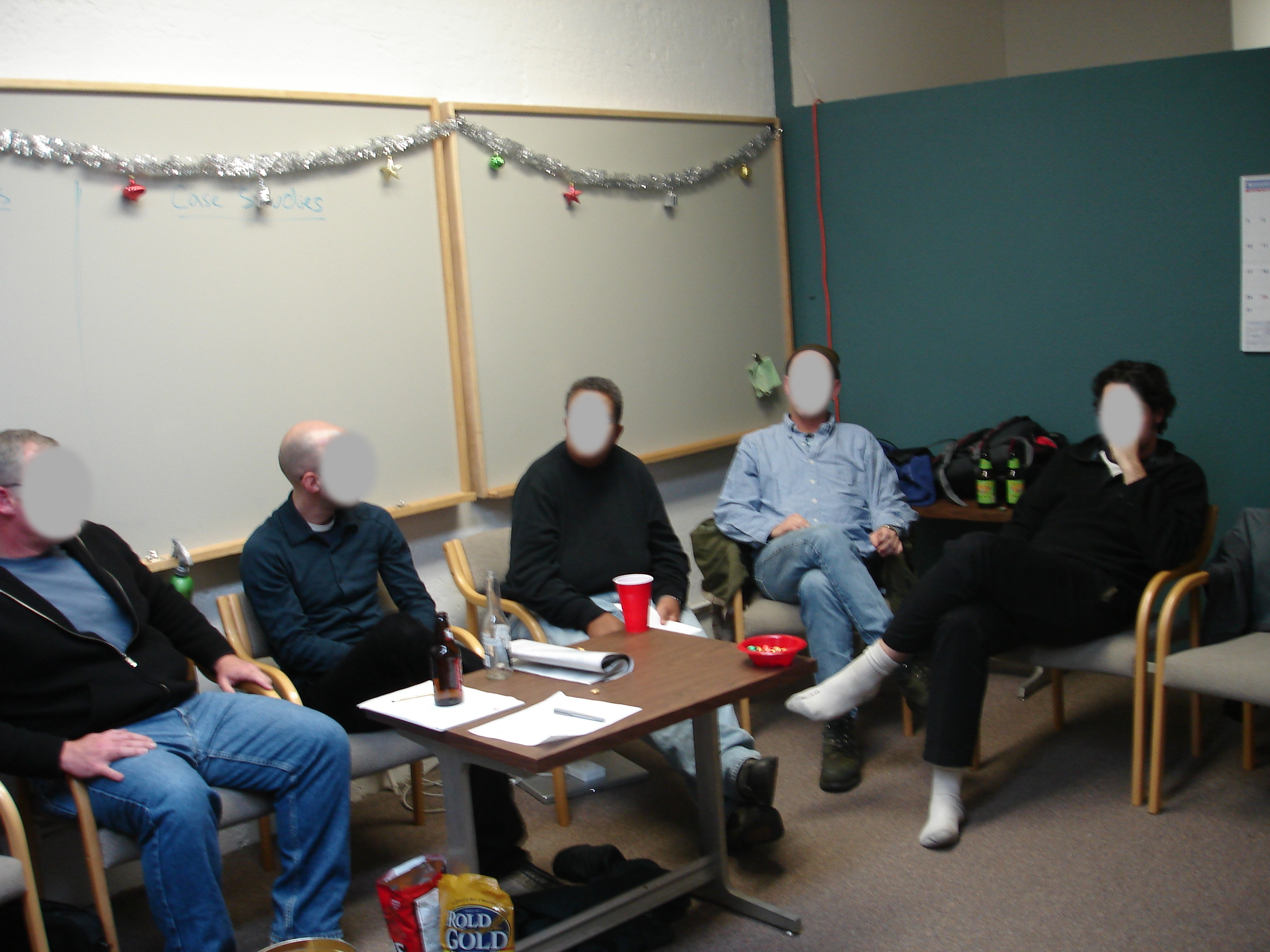} &
  \vspace{-5pt} 
  \scriptsize\selectfont{\textbf{Description}: The image depicts a group of six people sitting around a dining table in a classroom or office setting. They are engaged in a conversation, likely discussing something related to their work or studies... }
  
 \scriptsize \textbf{Extracted tags}: dining table, chairs, cups, bottle, backpacks, conversation, studies.  
  \\
\end{tabular}

\begin{tabular}{@{} m{0.3\columnwidth} m{0.6\columnwidth} @{}}
  \includegraphics[width=0.32\columnwidth]{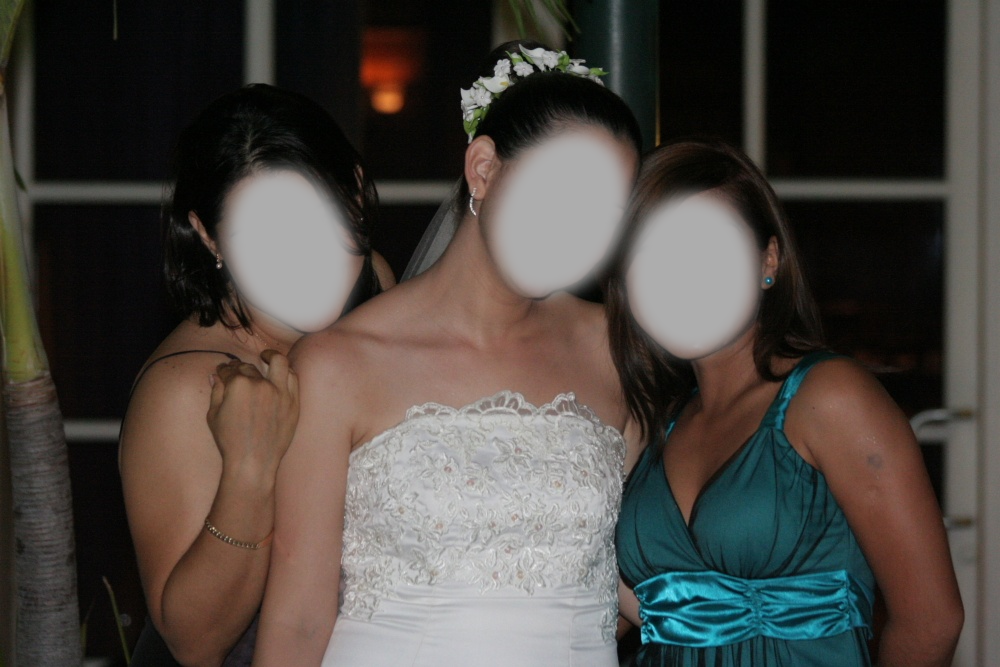} &
  \vspace{-5pt} 
  \scriptsize\selectfont{\textbf{Description}: The image features three women posing for a photo together. Two of the women are wearing wedding dresses, while the third woman is wearing a green dress. They are all smiling and appear to be enjoying each other's company... }
  
 \scriptsize \textbf{Extracted tags}: wedding dresses, green dress. 
  \\
\end{tabular}
\caption{DeX failure cases can occur when the LLM  doest not identify relevant keywords (e.g.~people). Such omissions highlight a potential bias in the dataset, where the mere presence of individuals is a sufficient condition for an image to be considered private. }
\label{fig:failure_cases}
\end{figure}
\noindent\textbf{Robustness}. We assess the predictive reliability against non-semantic perturbations by introducing zero-mean Gaussian noise to each image embedding. We consider two setups: 10  and 200 random vectors, denoted as $rand_{10}$ and $rand_{200}$, respectively. Figure~\ref{fig:random_perturbations} shows that random perturbations often cause prediction changes, but most flips occur with low confidence. 
At a confidence threshold of 0.5, the $V$ is 22\% for PrivacyAlert and 46\% for VISPR under $rand_{200}$, and drops to 3\% and 26\% when the confidence threshold is 0.7, respectively. 
\begin{figure*}[h!]
    \centering
    \includegraphics[width=0.9\columnwidth]{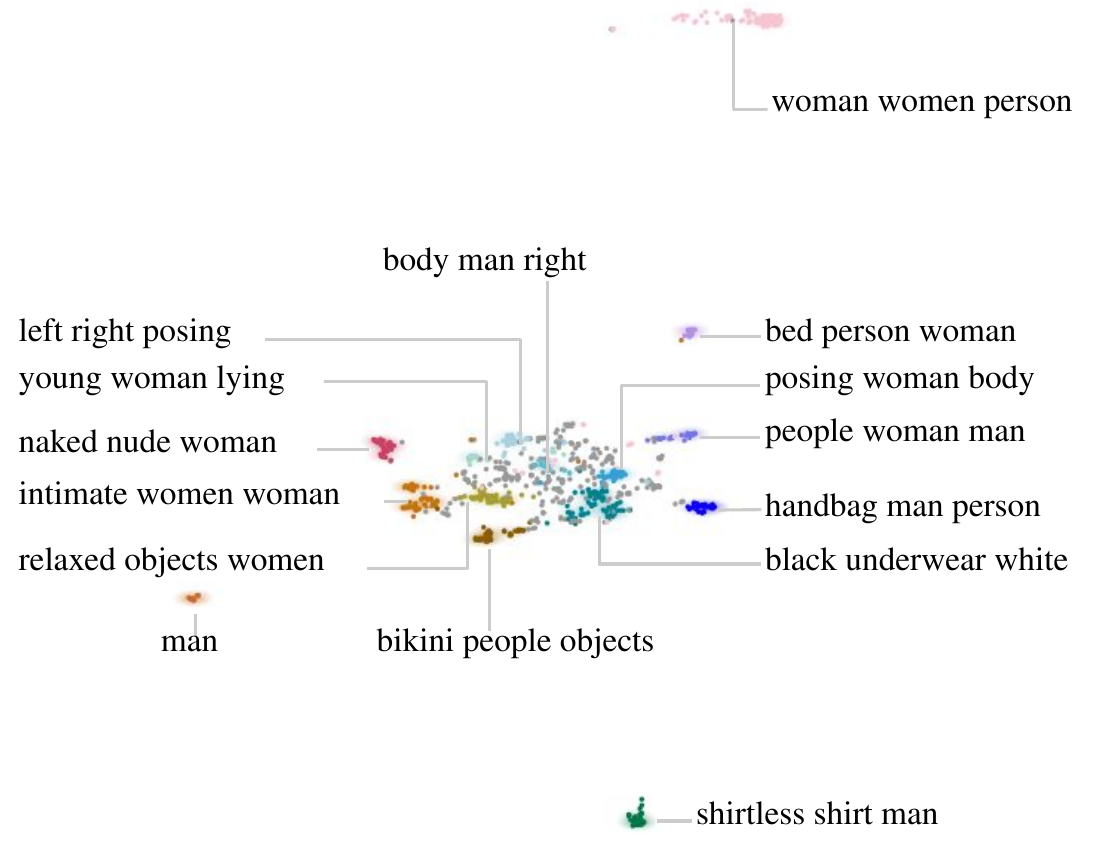}\hspace{10mm}    \includegraphics[width=1.05\columnwidth]{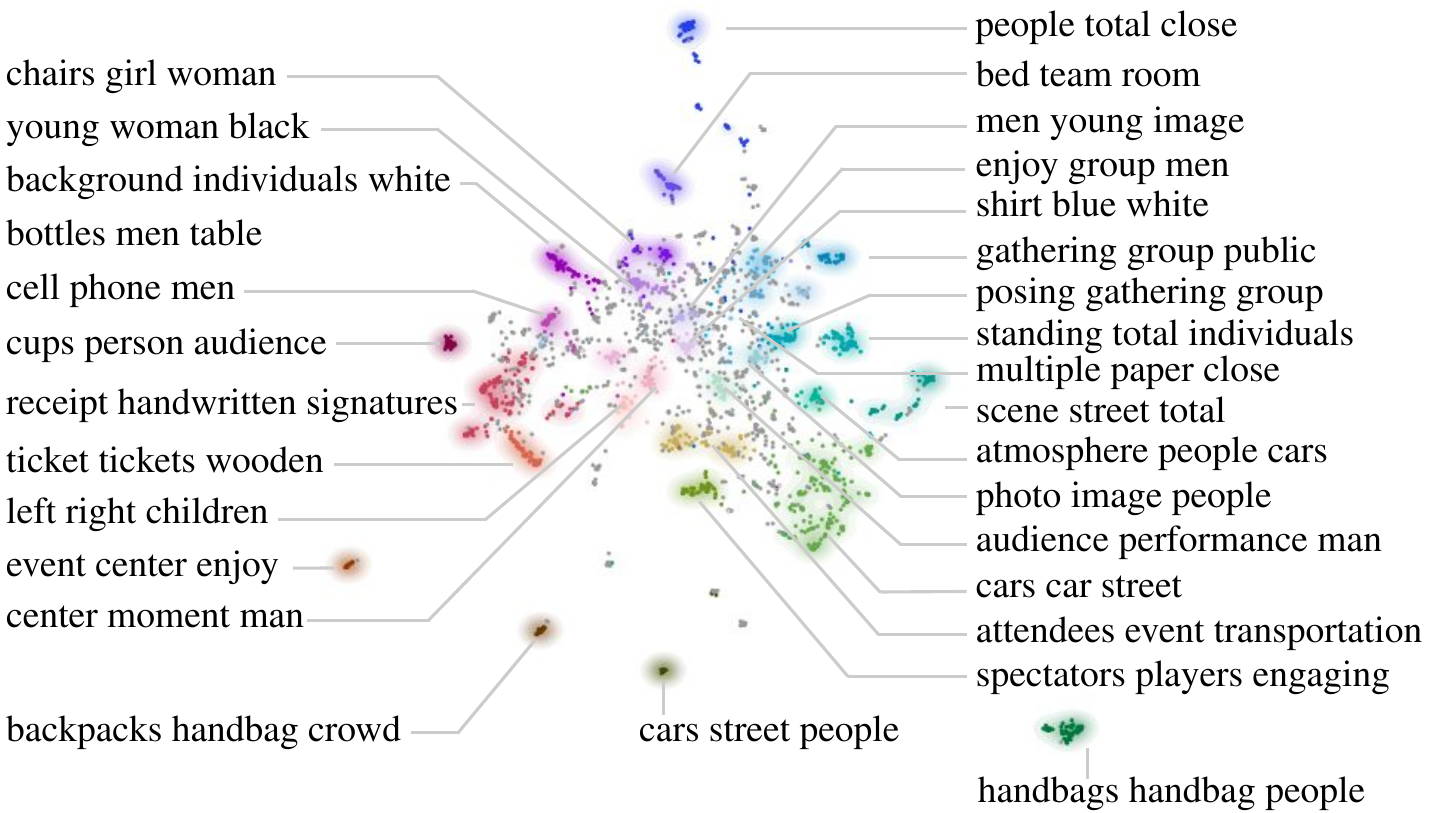}
    \caption{ List of topics discovered from the explanations generated on PrivacyAlert~\cite{PrivacyAlert} (left) and VISPR~\cite{orekondy_68_attributes} (right). In PrivacyAlert we observe that the main theme is nudity/sexual. In VISPR, different type of content is considered private, such as personal belongings (e.g., smartphones), cars, documents. Note that the topic names are generated automatically by the topic modeling algorithm~\cite{grootendorst2022bertopic}. }
    \label{fig:topics}
\end{figure*}

In contrast, concept-based manipulations lead to changes in predictions with higher confidence, and the $V$ remains high even when using a confidence threshold of 0.7.
We ensure meaningful explanations by filtering out predictions with low confidence.

\noindent\textbf{Failure cases}. DeX is bounded by the limitations of the tagging model, including VLM hallucinations. Failure cases for DeX might arise when the descriptions are too generic or imprecise, or when the LLM fails to extract relevant keywords (Figure~\ref{fig:failure_cases}).
For example, some images depicting scenes with people had accurate descriptions, but the LLM failed to extract the terms “people” or “person”. \\

\noindent\textbf{Identifying dataset bias}. DeX is usable outside the context of explainability and can also be used to analyze datasets and their biases. Given the explanations generated with DeX, we perform topic modeling to identify the main factors that influence the model's prediction and potential biases in the datasets.
We use BERTopic~\cite{grootendorst2022bertopic}. BERTopic uses HDBSCAN~\cite{hdbscan} to cluster the data using their embedding representation. Before clustering, the dimensionality of the data embeddings is reduced to 5 using UMAP~\cite{umap}, as HDBSCAN performs better on low-dimensional data. For visualization, the embeddings are reduced to 2 dimensions using UMAP. A hyperparameter search for finding the best topic modeling configuration and  managing synonyms and singular/plural forms in the topic names is out of scope for this work. Note that different setups might generate different results.

We identify significant differences between VISPR and PrivacyAlert: the former dataset has high diversity of private content and scenarios (i.e.~people at different events, cars with visible license plates,  documents, and personal objects like smartphones), whereas the private content of the latter is predominantly related to nudity/sexual material and intimate moments. The list of topics is shown in Figure~\ref{fig:topics}.

%%%%%%%%%%%%%%%%%%%%%%%%%%%%%%%%%%%%%%%%%%%%%%%%
\section{Conclusions}

We proposed DeX, a concept-based counterfactual cross-modal decompositionality method that produces text-based explanations. DeX uses image-specific concepts to modify image embeddings and identify the minimal set of concepts that influence the decision. By combining key image elements and evaluating multiple counterfactual scenarios, DeX produces multiple explanations per image, enabling a more comprehensive understanding of the model's decisions. DeX imposes sparsity by design, allowing the method to generate explanations using a fixed number of concepts for ease of interpretability. We applied DeX to the challenging case of image privacy and showed that it identifies the concepts that determine the classification decision. 
Moreover, DeX offers insights into dataset thematic biases, serving to enhance sample diversity.  Future work will focus on employing DeX in diverse privacy scenarios, such as predicting privacy scores via counterfactual concept manipulations to explain score sensitivity and in further tasks like fairness analysis. \\

\noindent\textbf{Acknowledgments.} Most of the research presented in this paper was conducted when the first author was affiliated with the Idiap Research Institute.

{
    \small
    \bibliographystyle{ieeenat_fullname}
    \bibliography{main}
}

\end{document}